\documentstyle[twocolumn]{article}
\title{Cues and control in Expert-Client Dialogues}
\author{Steve Whittaker \& Phil Stenton \\
	Hewlett-Packard Laboratories \\
	Filton Road, Bristol BS12 6QZ, UK. \\
	email: sjw@hplb.csnet
}
\date{}
\begin{document}
\maketitle

\begin{abstract}
          We conducted an empirical analysis into the relation between
          control and discourse structure. We applied control criteria to
          four dialogues and identified 3 levels of discourse structure. We
          investigated the mechanism for changing control between these
          structures and found that utterance type and not cue words
          predicted shifts of control. Participants used certain types of
          signals when discourse goals were proceeding successfully but
          resorted to interruptions when they were not.
\end{abstract}

\section{Introduction}

          A number of researchers have shown that there is organisation in
          discourse above the level of the individual utterance (5,
          8, 9, 10), The current exploratory study uses control as a
          parameter for identifying these higher level structures. We then
          go on to address how conversational participants co-ordinate
          moves between these higher level units, in particular looking at
          the ways they use to signal the beginning and end of such high
          level units.

          Previous research has identified three means by which speakers
          signal information about discourse structure to listeners: Cue
          words and phrases (5, 10); Intonation (7); Pronominalisation
	  (6, 2). In the cue words approach, Reichman
          (10) has claimed that phrases like ``because'', ``so'', and ``but''
          offer explicit information to listeners about how the speaker's
          current contribution to the discourse relates to what has gone
          previously. For example a speaker might use the expression ``so''
          to signal that s/he is about to conclude what s/he has just said.
          Grosz and Sidner (5) relate the use of such phrases to changes in
          attentional state. An example would be that ``and'' or ``but'' signal
          to the listener that a new topic and set of referents is being
          introduced whereas ``anyway'' and ``in any case'' indicate a return
          to a previous topic and referent set. A second indirect way of
          signalling discourse structure is intonation. Hirschberg and
          Pierrehumbert (7) showed that intonational contour is closely
          related to discourse segmentation with new topics being signalled
          by changes in intonational contour. A final more indirect cue to
          discourse structure is the speaker's choice of referring
          expressions and grammatical structure. A number of researchers
          (4, 2, 6, 10) have given accounts of how these relate to the
          continuing, retaining or shifting of focus.

          The above approaches have concentrated on particular surface
          linguistic phenomena and then investigated what a putative cue
          serves to signal in a number of dialogues. The problem with this
          approach is that the cue may only be an infrequent indicator of a
          particular type of shift. If we want to construct a general
          theory of discourse than we want to know about the whole range
          of cues serving this function. This study therefore takes a
          different approach. We begin by identifying all shifts of control
          in the dialogue and then look at how each shift was signalled by
          the speakers. A second problem with previous research is that the
          criteria for identifying discourse structure are not always made
          explicit. In this study explicit criteria are given: we then go
          on to analyse the relation between cues and this structure.

\section{The data}

          The data were recordings of telephone conversations between
          clients and an expert concerning problems with software. The tape
          recordings from four dialogues were then transcribed and the
          analysis conducted on the typewritten transcripts rather than the
          raw recordings. There was a total of 450 turns in the dialogues.

{\bf 2.1 Criteria for classifying utterance types.} Each utterance in the
dialogue was classified into one of four categories: (a)
          {\bf Assertions} - declarative utterances which were used to state
          facts. Yes or no answers to questions were also classified as
          assertions on the grounds that they were supplying the listener
          with factual information; (b) {\bf Commands} - utterances which were
          intended to instigate action in their audience. These included
          various utterances which did not have imperative form, (e.g.
          ``What I would do if I were you is to relink X'') but were intended
          to induce some action; (c) {\bf Questions} - utterances which were
          intended to elicit information from the audience. These included
          utterances which did not have interrogative form. e.g. ``So my
          question is....'' They also included paraphrases, in which the
          speaker reformulated or repeated part or all of what had just
          been said. Paraphrases were classified as questions on the
          grounds that the effect was to induce the listener to confirm or
          deny what had just been stated; (d) {\bf Prompts} - These were
          utterances which did not express propositional content. Examples
          of prompts were things like ``Yes'' and ``Uhu''.

{\bf 2.2 Allocation of control in the dialogues.}           We devised several
          rules to determine the location of control in the dialogues. Each
          of these rules related control to utterance type: (a) For
          questions, the speaker was defined as being in control unless the
          question directly followed a question or command by the other
          conversant. The reason for this is that questions uttered
          following questions or commands are normally attempts to clarify
          the preceding utterance and as such are elicited by the previous
          speaker's utterance rather than directing the conversation in
          their own right. (b) For assertions, the speaker was defined as
          being in control unless the assertion was made in response to a
          question, for the same reasons as those given for questions; an
          assertion which is a response to a question could not be said to
          be controlling the discourse; (c) For commands, the speaker was
          defined as controlling the conversation. Indirect commands (i.e.
          utterances which did not have imperative form but served to
          elicit some actions) were also classified in this way; (d) For
          prompts, the listener was defined as controlling the
          conversation, as the speaker was clearly abdicating his/her turn.
          In cases where a turn consisted of several utterances, the
          control rules were only applied to the final utterance.

          We applied the control rules and found that control did not
          alternate from speaker to speaker on a turn by turn basis, but
          that there were long sequences of turns in which control remained
          with one speaker. This seemed to suggest that the dialogues were
          organised above the level of individual turns into phases where
          control was located with one speaker. The mean number of turns in
          each phase was 8.03.

\section{Mechanisms for switching control}

          We then went on to analyse how control was exchanged between
          participants at the boundaries of these phases. We first examined
          the last utterance of each phase on the grounds that one
          mechanism for indicating the end of a phase would be for the
          speaker controlling the phase to give some cue that he (both
          participants in the dialogues were always male) no longer wished
          to control the discourse. There was a total of 56 shifts of
          control over the 4 dialogues and we identified 3 main classes of
          cues used to signal control shifts These were prompts,
          repetitions and summaries. We also looked at when no signal was
          given (interruptions).

{\bf 3.1 Prompts.} On 21 of the 56 shifts (38\%), the utterance
          immediately prior to the control shift was a prompt. We might
          therefore explain these shifts as resulting from the person in
          control explicitly indicating that he had nothing more to say.

          (In the following examples a line indicates a control
          shift)

          Example 1 - Prompt  Dialogue C -

\begin{quote}
1. E: "And they are, in your gen you'll find that they've
          relocated into the labelled common area" (E control)\\

2. C: "That's right." (E control)\\

3. E: "Yeah" (E abdicates control with prompt)\\
-----------------------------\\
4. C: "I've got two in there. There are two of them." (C control)\\

5. E: "Right" (C control)\\

6. C: "And there's another one which is \% RESA" (C control)\\

7. E: "OK um" (C control)\\

8. C: "VS" (C control)\\

9. E: "Right" (C control)\\

10. C: "Mm" (C abdicates control with prompt)\\
----------------------------\\
11. E: "Right and you haven't got - I assume you haven't got
          local labelled common with those labels" (E control)\\
\end{quote}

{\bf 3.2 Repetitions and summaries.} On a further 15 occasions (27\%),
          we found that the person in control of the dialogue signalled
          that they had no new information to offer. They did this either
          by repeating what had just been said (6 occasions), or by giving
          a summary of what they had said in the preceding utterances of
          the phase (9 occasions). We defined a repetition as an assertion
          which expresses part or all of the propositional content of a
          previous assertion but which contains no new information. A
          summary consisted of concise reference to the entire set of
          information given about the client's problem or the solution
          plan.

          Example 2 - Repetition. Dialogue C -

\begin{quote}
1. Client: "These routines are filed as DS" (C control)\\

2. Expert: "That's right, yes" (C control)\\

3. C: "DS" (C abdicates control with repetition)\\
--------------------------------\\
4. E: "And they are, in your gen you'll find they've relocated

          into your local common area." (E control)\\
\end{quote}

          Half the repetitions were accompanied by cue words. These were
          ``and'', ``well'' and ``so'', which prefixed the assertion.

          Example 3 - Summary Dialogue B -
\begin{quote}
1. E. "OK. Initialise the disc retaining spares" (E control)\\

2. C: "Right" (E control)\\

3. E: "Uh and then TF it back" (E control)\\

4. C: "Right" (E control)\\

5. E: "Did you do the TF with verify?" (E control)\\

6. C: "Er yes I did" (E control)\\

7. E: "OK. That would be my recommendation and that will ensure
          that you get er a logically integral set of files" (E abdicates
          control with summary)\\
-------------------------------\\
8. C: "Right. You think that initialising it using this um EXER
          facility." (C control)\\
\end{quote}

          What are the linguistic characteristics of summaries? Reichman
          (10) suggests that ``so'' might be a summary cue on the part of the
          speaker but we found only one example of this, although there
          were 3 instances of ``and'', one ``now'' one ``but'' and one ``so''.
In
          our dialogues the summaries seemed to be characterised by the
          concise reference to objects or entities which had earlier been
          described in detail, e.g. (a) ``Now, I'm wondering how the two are
          related'' in which ``the two'' refers to the two error messages
          which it had taken several utterances to describe previously. The
          other characteristic of summaries is that they contrast strongly
          with the extremely concrete descriptions elsewhere in the
          dialogues, e.g. ``err the system program standard call file
          doesn't complete this means that the file does not have a tail
          record'' followed by ``And I've no clue at all how to get out of
          the situation''. Example 3 also illustrates this change from
          specific (1, 3, 5) to general (7).          How then do repetitions
and summaries operate as cues?  In
          summarising, the speaker is indicating a natural breakpoint in
          the dialogue and they also indicate that they have nothing more
          to add at that stage. Repetitions seem to work in a similar way:
          the fact that a speaker reiterates indicates that he has nothing
          more to say on a topic.

{\bf 3.3 Interruptions.} In the previous cases, the person controlling
          the dialogue gave a signal that control might be exchanged. There
          were 20 further occasions (36\% of shifts) on which no such
          indication is given. We therefore went on to analyse the
          conditions in which such interruptions occurred. These seem to
          fall into 3 categories: (a) vital facts; (b) responses to vital
          facts; (c) clarifications.

{\bf 3.3.1 Vital facts.} On a total of 6 occasions (11\% of shifts) the
          client interrupted to contradict the speaker or to supply what
          seemed to be relevant information that he believed the expert did
          not know.

          Example 4  Dialogue C -
\begin{quote}
1. E: ".... and it generates this warning, which is now at 4.0 to
          warn you about the situation"  (E control)\\
---------------------------------\\
2. C: "It is something new though um" (C assumes control by
          interruption)\\

3. E: "Well" (C control)\\

4. C: "The programs that I've run before obviously LINK A's got
          some new features in it which er..." (C control)\\
---------------------------------\\
5. E: "That's right, it's a new warning at 4.0" (E assumes
    control by interruption)\\
\end{quote}
          Two of these 6 interjections were to supply extra information and
          one was marked with the cue ``as well''. The other four were to
          contradict what had just been said and two had explicit markers
          ``though'' and ``well actually'': the remaining two being direct
          denials.

{\bf 3.3.2 Reversions of control following vital facts.} The next
          class of interruptions occur after the client has made some
          interjection to supply a missing fact or when the client has
          blocked a plan or rejected an explanation that the expert has
          produced. There were 8 such occasions (14\% of shifts).

          The interruption in the previous example illustrates the
          reversion of control to the expert after the client has supplied
          information which he (the client) believes to be highly relevant
          to the expert. In the following example, the client is already in
          control.\newline

          Example 5  Dialogue B -
\begin{quote}
1. "I'll take a backup first as you say" (C control)\\

2. E: "OK" (C control)\\

3. C: "The trouble is that it takes a long time doing all this"
          (C control)\\
------------------------------------\\
4. E: "Yeah, yeah but er this kind of thing there's no point
          taking any short cuts or you could end up with no system at all."
          (E assumes control by interruption)\\

          On five occasions the expert explicitly signified his acceptance
          or rejection of what the client had said, e.g.``Ah'',``Right
          indeed'',``that's right'',``No'',``Yeah but''. On three occasions
          there were no markers.
\end{quote}
{\bf 3.3.3 Clarifications.} Participants can also interrupt to clarify
          what has just been said. This happened on 6 occasions (11\%) of
          shifts.

          Example 6  Dialogue C -
\begin{quote}
1. C: "If I put an SE in and then do an EN it comes up"  (C
          control)\\
----------------------------------\\
2. E: "So if you put in a ...?" ( E control)\\

3. C: "SE" (E control)\\
\end{quote}
          On two occasions clarifications were prefixed by ``now'' and twice
          by ``so''.  On the final two occasions there was no such marker,
          and a direct question was used.

{\bf 3.3.4 An explanation of interruptions.} We have just described the
          circumstances in which interruptions occur, but can we now
          explain why they occur? We suggest the following two principles
          might account for interruptions: these principles concern: (a)
          the information upon which the participants are basing their
          plans, and (b) the plans themselves.
\begin{description}
\item[(A). Information quality:]          Both expert and client must believe
that the information that the
          expert has about the problem is true and that this information is
          sufficient to solve the problem. This can be expressed by the
          following two rules which concern the truth of the information
          and the ambiguity of the information: (A1) if the listener
          believes a fact P and believes that fact to be relevant and
          either believes that the speaker believes not P or that the
          speaker does not know P then interrupt; (A2) If the listener
          believes that the speaker's assertion is relevant but ambiguous
          then interrupt.
\item[(B). Plan quality:] Both expert and client must
          believe that the plan that the expert has generated is adequate
          to solve the problem and it must be comprehensible to the client.
          The two rules which express this principle concern the
          effectiveness of the plan and the ambiguity of the plan: (B1) If
          the listener believes P and either believes that P presents an
          obstacle to the proposed plan or believes that part of the
          proposed plan has already been satisfied, then interrupt; (B2) If
          the listener believes that an assertion about the proposed plan
          is ambiguous, then interrupt.
\end{description}

In this framework, interruptions
          can be seen as strategies produced by either conversational
          participant when they perceive that a either principle is not
          being adhered to.

{\bf 3.4 Cue reliability.} We also investigated whether there were
          occasions when prompts, repetitions and summaries failed to
          elicit the control shifts we predicted. We considered two
          possible types of failure: either the speaker could give a cue
          and continue or the speaker could give a cue and the listener
          fail to respond. We found no instances of the first case;
          although speakers did produce phrases like ``OK'' and then
          continue, the ``OK'' was always part of the same intonational
          contour as that further information and there was no break
          between the two, suggesting the phrase was a prefix and not a
          cue. We did, however, find instances of the second case: twice
          following prompts and once following a summary, there was a long
          pause, indicating that the speaker was not ready to respond. We
          conducted a similar analysis for those cue words that have been
          identified in the literature. Only 21 of the 35 repetitions,
          summaries and interruptions had cue words associated with them
          and there were also 19 instances of the cue words ``now'', ``and'',
          ``so'', ``but'' and ``well'' occurring without a control shift.

\section{Control cues and global control}

          The analysis so far has been concerned with control shifts where
          shifts were identified from a series of rules which related
          utterance type and control. Examination of the dialogues
          indicated that there seemed to be different types of control
          shifts: after some shifts there seemed to be a change of topic,
          whereas for others the topic remained the same. We next went on
          to examine the relationship between topic shift and the different
          types of cues and interruptions described earlier. To do this it
          was necessary first to classify control shifts according to
          whether they resulted in shifts of topic.

{\bf 4.1 Identifying topic shifts.} We identified topic shifts in the
          following way: Five judges were presented with the four dialogues
          and in each of the dialogues we had marked where control shifts
          occurred. The judges were asked to state for each control shift
          whether it was accompanied by a topic shift. All five judges
          agreed on 24 of the 56 shifts, and 4 agreed for another 22 of the
          shifts. Where there was disagreement, the majority judgment was
          taken.

{\bf 4.2 Topic shift and type of control shift.}  Analysing each type of
          control shift, it is clear that there are differences between the
          cues used for the topic shift and the no shift cases. For
          interruptions, 90\% occur within topic, i.e. they do not result in
          topic shifts. The pattern is not as obvious for prompts and
          repetitions/summaries, with 57\% of prompts occurring within topic
          and 67\% of repetitions/summaries occurring within topic. This
          suggests that change of topic is a carefully negotiated process.
          The controlling participant signals that he is ready to close the
          topic by producing either a prompt or a repetition/summary and
          this may or may not be accepted by the other participant. What is
          apparent is that it is highly unusual for a participant to seize
          control and change topic by interruption. It seems that on the
          majority of occasions (63\%) participants wait for the strongest
          possible cue (the prompt) before changing topic.

{\bf 4.3 Other relations between topic and control.} We also looked at
          more general aspects of control within and between topics. We
          investigated the number of utterances for which each participant
          was in control and found that there seemed to be organisation in
          the dialogues above the level of topic. We found that each
          dialogue could be divided into two parts separated by a topic
          shift which we labelled the central shift. The two parts of the
          dialogue were very different in terms of who controlled and
          initiated each topic. Before the central shift, the client had
          control for more turns per topic and after it, the expert had
          control for more turns per topic. The respective numbers of turns
          client and expert are in control before and after the central
          shift are :Before 11-7,22-8,12-6,21-6; After 12-33,16-23,2-11,0-5
for the four dialogues.
          With the exception of the first topic in Dialogues 1 and 4, the
          client has control of more turns in every topic before the
          central shift, whereas after it, the expert has control for more
          turns in every topic. In addition we looked at who initiated each
          topic, i.e. who produced the first utterance of each topic. We
          found that in each dialogue, the client initiates all the topics
          before the central shift, whereas the expert initiates the later
          ones. We also discovered a close relationship between topic
          initiation and topic dominance. In 19 of the 21 topics, the
          person who initiated the topic also had control of more turns. As
          we might expect, the point at which the expert begins to have
          control over more turns per topic is also the point at which the
          expert begins to initiate new topics.

\section{Conclusions}

          The main result of this exploratory study is the finding that
          control is a useful parameter for identifying discourse
          structure. Using this parameter we identified three levels of
          structure in the dialogues: (a) control phases; (b) topic; and
          (c) global organisation. For the control phases, we found that
          three types of utterances (prompts, repetitions and summaries)
          were consistently used to signal control shifts. For the low
          level structures we identified, (i.e. control phases), cue words
          and phrases were not as reliable in predicting shifts.  This
          result challenges the claims of recent discourse theories (5, 10)
          which argue for a the close relation between cue words and
          discourse structure. We also examined how utterance type related
          to topic shift and found that few interruptions introduced a new
          topic. Finally there was evidence for high level structures in
          these dialogues as evidenced by topic initiation and control,
          with early topics being initiated and dominated by the client and
          the opposite being true for the later parts.

          Another focus of current research has been the modelling of
          speaker and listener goals (1, 3) but there has been little
          research on real dialogues investigating how goals are
          communicated and inferred. This study identifies surface
          linguistic phenomena which reflect the fact that participants are
          continuously monitoring their goals. When plans are perceived as
          succeeding, participants use explicit cues such as prompts,
          repetitions and summaries to signal their readiness to move to
          the next stage of the plan. In other cases, where participants
          perceive obstacles to their goals being achieved, they resort to
          interruptions and we have tried to make explicit the rules by
          which they do this.

          In addition our methodology is different from other studies
          because we have attempted to provide an explanation for whole
          dialogues rather than fragments of dialogues, and used explicit
          criteria in a bottom-up manner to identify discourse structures.
          The number of dialogues was small and taken from a single problem
          domain. It seems likely therefore that some of our findings (e.g
          the central shift) will be specific to the diagnostic dialogues
          we studied. Further research applying the same techniques to a
          broader set of data should establish the generality of the
          control rules suggested here.

\end{document}